\documentclass[runningheads]{llncs}

\usepackage[T1]{fontenc}
\usepackage[utf8]{inputenc}
\usepackage{microtype}
\usepackage{amsmath}
\usepackage{amssymb}
\usepackage{inconsolata}

\usepackage{graphicx}
\usepackage{booktabs}
\usepackage{tabularx}
\usepackage{array}
\usepackage{xcolor}
\usepackage{hyperref}

\usepackage[square,numbers,sort&compress]{natbib}
\usepackage{enumitem}
\setlist[itemize]{leftmargin=1.2em, topsep=2pt, partopsep=0pt, parsep=0pt, itemsep=1pt}
\setlist[enumerate]{leftmargin=1.2em, topsep=2pt, partopsep=0pt, parsep=0pt, itemsep=1pt}

\hypersetup{
    colorlinks=true,
    linkcolor=blue,
    citecolor=blue,
    urlcolor=blue
}

\newcommand{\code}[1]{\texttt{#1}}

\begin{document}

\title{Unfragmenting Nawatl NMT: Single-Variant Specialization for Masewalkopa}
\subtitle{LoRA Adaptation, Morphosyntactic Conditioning, and Decolonial Evaluation}
\titlerunning{Single-Variant Specialization for Masewalkopa}

\author{Luis Armando Mercado-Campos\thanks{Corresponding author.} \and
Alexis Palmer}

\authorrunning{Luis A. Mercado-Campos and A. Palmer}

\institute{Department of Linguistics, University of Colorado Boulder\\
\email{Luis.MercadoCampos@colorado.edu}, \email{alexis.palmer@colorado.edu}}

\maketitle

\begin{abstract}
In minoritized (low-resource) language technologies, a common response to data sparsity is to aggregate text across multiple varieties and treat the combined data as a single corpus. However, mixed corpora of closely related indigenous variants can produce an artificial register that is accountable to no single speaker community, a structural failure we call \textit{variant fragmentation}. This issue is acute for minoritized languages native to Mexico, such as Nawatl. Building on a 70,000-pair multi-variant foundation~\citep{mercadocampos2023design}, we specialize an mT5-large checkpoint to Masewalkopa (nhn), the Nawatl variety spoken in Santa Ana Tlacotenco. We curate 2,665 parallel pairs, reserve all sentence-level data for evaluation, and incorporate 192 elder-transcribed target sentences via asymmetric backtranslation. A custom finite-state analyzer reduces unanalyzed surface forms from 82.0\% to 19.9\% and supplies functional syntax tags as plain-text prefixes, allowing the model to reuse existing subword embeddings rather than relying on frozen special tokens. The resulting \texttt{lora\_v9c\_b} model achieves 58.46 dev chrF++ and 58.20/57.90 directional test chrF++. Finally, CTranslate2 int8 quantization reduces CPU latency from 137.5 to 1.8--2.6 seconds per request, enabling cost-bounded Cloud Run deployment.
\hfill\break

\keywords{Neural Machine Translation \and Minoritized NLP \and Nawatl \and Masewalkopa \and LoRA \and Finite-State Morphology \and Decolonial Evaluation.}
\end{abstract}

\section{Introduction and Linguistic Background}
\label{section:intro}

Nawatl is a group of endangered, closely-related, morphologically rich languages\footnote{I use ``language,'' ``variant,'' and ``variety'' interchangeably, but avoid the use of the term ``dialect.'' In Mexico, this distinction carries real social, educational, and political consequences, as historically the term ``dialect'' has been used as a tool to marginalize indigenous language speakers \citep{aguayo2016expressions}. In government documentation, schooling, and public discourse, referring to these languages as ``dialects'' can contribute to marginalization and under-resourcing \citep{parodi2011multiglosia}.} in the Nawan branch of the Uto-Aztecan family, spoken primarily in Mexico. There are 30 formally recognized variants of Nawatl \citep{inali:lenguas}, which can have substantial phonological, morphological, syntactic, and lexical differences \citep{gruda2023lexical}. There is also significant variant-internal diversity. Treating them as a single computational language obscures the systems that index place and community belonging.

Masewalkopa is the endonym used by speakers of the Nawatl in Santa Ana de Tlacotenco, Milpa Alta, Mexico City (\textit{nhn}). It is morphologically rich and predominantly head marking: words can encode subject, object, derivation, aspect, possession, predicate-nominal structure, and honorific morphology. Character-level similarity can therefore conceal a wrong possessor or a form from another writing tradition.

During the first author's three years of studying Masewalkopa, he encountered persistent discrepancies between locally taught forms and Classical Nawatl-centered digital resources. As a result, his objective shifted from maximizing a generic ``Nawatl'' score to producing forms that can be inspected by a single, identifiable community.

\subsection{From a Multi-Variant Model to Single-Variant Specialization}
\label{subsection:multi-variant-thesis-model-to-specialization}

In the first author's previous work~\citep{mercadocampos2023design}, he assembled approximately 70,000 Spanish-Nawatl pairs across varieties and domains. The resulting system (found on huggingface\footnote{https://huggingface.co/luisarmando/mt5-large-es-nah}) checkpoint achieved 54.15 chrF++ on a split from that same mixed collection.


However, when evaluated zero-shot on an external benchmark, the AmericasNLP 2021 Shared Task on Open Machine Translation for Indigenous Languages~\citep{americasnlp2021}, performance collapsed to 25.7 chrF++. This external evaluation dataset consisted of independent Spanish-Nawatl text that was completely unseen during training. Beyond simple domain mismatch, manual inspection revealed that the model generated conflicting lexical, orthographic, and inflectional forms from different regions within the same sentence. While each piece might be attested somewhere in Mexico, the combined output was incoherent as Masewalkopa.

We use \emph{variant fragmentation} for this interaction between data aggregation and model behavior. The multi-variant corpus expresses a single underlying meaning through divergent regional realizations, the subword model shares pieces regardless of regional provenance, and beam search can assemble a fluent-looking sequence from conflicting varietal conventions. Aggregate evaluation conceals the problem when references come from the same mixture. A high in-domain score can therefore coexist with low external validity.

To resolve this, we warm-start from the first author's 1.2-billion-parameter multi-variant \texttt{mT5-large} thesis checkpoint and adapt it exclusively with Masewalkopa data. The broad-base model provides sequence modeling and cross-lingual alignment, while the small community corpus provides regional constraints. Low-Rank Adaptation (LoRA) makes this transfer feasible on a single GPU via compact, low-rank updates~\citep{hu2022lora}.

In this paper, we develop a single-variety approach to neural machine translation (NMT) for Masewalkopa, using a relatively small community dataset to specialize an existing multi-variety model. By combining parameter-efficient LoRA fine-tuning with rule-based finite-state morphology and authentic target-side oral narratives, we show that targeted specialization resolves variant fragmentation while respecting the linguistic autonomy of Santa Ana Tlacotenco speech.

\subsection{Core Contributions}
\label{subsection:core-contributions}

This paper makes four contributions:
\begin{enumerate}
    \item \textbf{Single-Variant Specialization Pipeline:} We specialize a 1.2-billion-parameter NMT model with 2,665 Masewalkopa pairs, restricting short dictionary entries to training so that development and test benchmarks evaluate only full sentence translation.
    \item \textbf{Morphosyntactic NMT Conditioning:} We expand a finite-state analyzer beyond 6,500 lexicon lines to derive surface-validated functional syntax prefixes, supplying the neural translation encoder with structural grounding to resolve morphological and lexical ambiguities.
    \item \textbf{Ablation Suite and Post-Mortems:} We ablate asymmetric backtranslation, cold special tokens, sparse versus dense syntax, and syntax stripping in the final \code{lora\_v9c} experiment.
    \item \textbf{Decolonial and Production Evaluation:} We combine chrF++, linguistic diagnostics, DBS, human comparison, and measured CPU deployment behavior.
\end{enumerate}

Together, these experiments demonstrate that targeted single-variant specialization overcomes the pitfalls of multi-variant data aggregation. The final specialized model (\texttt{lora\_v9c\_b}) achieves 58.46 dev chrF++ and 58.20/57.90 directional test chrF++, gaining over 28 points over the mixed-variant base. Ablation results show that plain-text syntactic prefixes successfully recover from the severe optimization stalls of frozen special tokens, while dense morphological tagging yields measurable structural gains (+1.33 test chrF++). Culturally, decolonial DBS probes and human A/B ratings confirm a strong reduction in colonial Spanish substitutions and improved naturalness on elder-transcribed oral narratives. Finally, 8-bit quantization and widened beam search reduce inference latency to under 2.6 seconds on standard CPU hardware, demonstrating that single-variant models can be operated within strict community budget constraints.

\section{Related Work}
\label{sec:related_work}

\paragraph{Indigenous Machine Translation and Variety Specialization.}
Prior machine translation research on Indigenous languages of the Americas has largely focused on pooling heterogeneous regional variants into aggregate training sets to combat extreme data scarcity~\citep{feldman2020neural}. However, computational dialectology and variety classification for Nawatl demonstrates that regional variants exhibit sharp lexical and morphophonological boundaries~\citep{mercado-campos-etal-2026-power}. Blending these divergent systems into monolithic training corpora frequently produces varietal bleed and privileges colonial or standardized varieties over living community speech~\citep{bird2020decolonising}. Recent work therefore emphasizes community-grounded single-variant modeling. This study builds on that foundation by specializing a broad multi-variant base directly to the spoken variety of Santa Ana Tlacotenco.

\paragraph{Morphosyntactic Conditioning in Neural MT.}
Guiding sequence generation with explicit linguistic structure stems from factored translation architectures~\citep{sennrich2016linguistic} and prefix-based syntactic conditioning~\citep{li2020syntax}. For polysynthetic languages, finite-state transducers (FSTs) supply robust morphological segmentation and grammatical disambiguation where purely statistical subword models fail~\citep{beesley2003finite}. My system builds on these insights, demonstrating that high-density FST syntactic prefixes provide reliable structural inductive bias without requiring vocabulary modifications.

\paragraph{Parameter-Efficient Adaptation in Low-Resource Regimes.}
Low-Rank Adaptation (LoRA)~\citep{hu2022lora} provides an efficient mechanism for adapting massive multilingual models such as \code{mT5}~\citep{xue2021} while preventing catastrophic forgetting. Extending low-rank updates across all linear projections is essential for deep domain transfer~\citep{dettmers2023qlora, he2022unified}. We contribute an empirical analysis of embedding constraints under LoRA, showing how plain-text subword formatting avoids the optimization collapse associated with frozen cold-token representations.
\section{Data Curation, Corpus Composition, and Governance}

We compiled the primary parallel corpus over nearly four years of studying Masewalkopa through community classes and structured coursework with native-speaker language instructors from Santa Ana Tlacotenco. All pedagogical exercises, classroom dialogues, and vocabulary transcriptions were recorded, digitized, and curated with the explicit informed consent of the instructors for research and educational work.
Following the Collective Benefit, Authority to COntrol, Responsibility, and Ethics (CARE) principles, we treat provenance, authority, responsibility, and ethics as pipeline properties~\citep{carroll2020}. Data collection was guided by collective benefit and mutual accountability rather than extractive aggregation.

\subsection{Cleaning and Provenance}
\label{subsection:cleaning-provenance}

In machine translation, the main training unit is an aligned parallel pair consisting of a source text segment and its equivalent target translation. We compiled the primary corpus over nearly four years of studying Masewalkopa through classes and coursework with native-speaker instructors from Santa Ana Tlacotenco. All pedagogical exercises, classroom dialogues, and vocabulary notes were recorded and curated into structured spreadsheets with explicit instructor consent.

We normalized whitespace and Unicode representations, removed empty fields, and deduplicated aligned Spanish--Nawatl segment pairs case-insensitively. Near-neighbor spellings were retained because they often encode authentic local phonological choices. Cleaning reduced 2,761 raw rows to 2,665 unique parallel segment pairs.

Each parallel pair was tagged by sentential status based on word count from standard whitespace splitting after punctuation stripping. Entries with at most two whitespace-delimited tokens were treated as lexical or short-phrase material (such as citation nouns or isolated verb stems); longer examples and independently verified clauses were treated as sentence-level. This reproducible threshold does not claim that every two-token sequence lacks syntax. Instead, it prevents dictionary correspondences from dominating evaluation splits.

\begin{table}[!htbp]
\caption{Composition of the cleaned parallel corpus before bidirectional expansion, with representative examples.}
\label{tab:corpus_composition}
\centering
\small
\resizebox{\textwidth}{!}{%
\begin{tabular}{lp{1.1cm}p{1.0cm}p{3.8cm}p{3.8cm}}
\toprule
\textbf{Subset} & \textbf{Pairs} & \textbf{Share} & \textbf{Illustrative Examples} & \textbf{Primary Experimental Role} \\
\midrule
Lexical / short ($\le 2$ tokens) & 1,188 & 44.6\% & \textit{sesek} $\leftrightarrow$ \textit{frío}; \newline \textit{nimitstlasohkamate} $\leftrightarrow$ \textit{te agradezco} & Training coverage, citation forms, short-form deployment probes \\
\addlinespace[4pt]
Sentence level & 1,477 & 55.4\% & \textit{tlamach tikayawaskeh} $\leftrightarrow$ \textit{juntos en comunidad venceremos} & Training plus exclusive source of development and test items \\
\midrule
\textbf{Total unique} & \textbf{2,665} & \textbf{100.0\%} & \multicolumn{2}{l}{Masewalkopa specialization corpus} \\
\bottomrule
\end{tabular}}
\end{table}

Nearly 45\% of the pairs are lexical or very short. They teach stems and spelling but underrepresent agreement, possession, negation, and discourse. Random splitting would therefore reward dictionary matches without demonstrating sentence translation.

\subsection{Stratification and Bidirectional Tasks}
\label{subsection:stratification-bidirectional-tasks}

Development and test data were drawn exclusively from the 1,477 sentence-level pairs to ensure that evaluation benchmarks measure genuine sequence translation rather than isolated vocabulary lookup. Including single-word or short lexical pairs in held-out sets would artificially inflate n-gram and character overlap scores through simple dictionary matching, masking failures in clause-level syntax, agreement, and possession. Restricting evaluation to sentential items guarantees that metrics reflect grammatical compositionality, while placing all 1,188 lexical pairs into the training inventory provides necessary stem coverage and orthographic grounding.

Using a fixed stratification seed, we reserved 120 sentence pairs for development and 150 for the final held-out test set. The remaining training inventory contained 2,335 pairs after recombining the lexical entries with eligible training sentences. 

To train the model bidirectionally, each parallel sentence pair consisting of a Spanish segment $(s_{\text{es}})$ and a Masewalkopa segment $(s_{\text{nah}})$ was expanded into two separate supervised training examples by prepending a literal directional prefix to the source string:
\begin{itemize}
    \item \textbf{Spanish $\rightarrow$ Masewalkopa:} \\
    \textit{Input string:} \texttt{es->nah: juntos venceremos como comunidad} \\
    \textit{Target string:} \texttt{tlamach tikayawaskeh}
    \item \textbf{Masewalkopa $\rightarrow$ Spanish:} \\
    \textit{Input string:} \texttt{nah->es: tlamach tikayawaskeh} \\
    \textit{Target string:} \texttt{juntos venceremos como comunidad}
\end{itemize}

The 2,335 training pairs therefore yielded 4,670 directional training examples, while the 120 development pairs and 150 test pairs yielded 240 and 300 directional evaluation examples, respectively. 

Directional prefixes are ordinary text already present in the mT5 SentencePiece vocabulary. This detail is crucial because standard text prefixes remain trainable through attention and LoRA weight updates even when the base model's shared embedding table is completely frozen. Joint bidirectional training doubles the available supervised training instances without requiring synthetic target generation.

\subsection{Four Community Oral Texts}
\label{subsection:four-community-oral-texts}

To add connected speech and local cultural vocabulary, we incorporated 192 elder-transcribed Masewalkopa sentences. These materials were generously shared with me for pedagogical development by one of my native speaker language instructors in Santa Ana Tlacotenco. The texts represent four distinct thematic registers:

\begin{enumerate}
    \item \textbf{Tiankisko:} 34 market-dialogue sentences covering exchanges, quantities, and polite requests.
    \item \textbf{Tonakayutl:} 45 sentences on traditional maize cultivation, harvesting practices, and food knowledge.
    \item \textbf{Kowatl itech in ikwentsin:} 50 narrative sentences describing a traditional snake story, adding complex reference tracking and sequential event structures.
    \item \textbf{Tatahpowilistle:} 63 sentences concerning community authority, local oral history, and traditional institutions.
\end{enumerate}

Preserving these document titles and categories respects their local genre provenance, even though the current translation model does not condition directly on document identity.

\subsection{Asymmetric Target-Side Backtranslation}
\label{subsection:asymmetric-target-side-backtranslation}

Backtranslation generates synthetic sources from monolingual target data~\citep{sennrich2016}. We used \code{lora\_v2} (45.58 Masewalkopa-to-Spanish test chrF++) to translate the 192 authentic Masewalkopa sentences into Spanish:

\begin{equation}
\hat{x}_{\text{es}} = \arg\max_{x} P_{\texttt{lora\_v2}}(x \mid y_{\text{nah}}^{\text{human}}),
\qquad
(\hat{x}_{\text{es}},\, y_{\text{nah}}^{\text{human}}).
\label{eq:bt}
\end{equation}

Unlike the curated pairs in Section~\ref{subsection:stratification-bidirectional-tasks}, which were expanded bidirectionally, these backtranslated pairs were used in one direction only ($\text{nah}\rightarrow\text{es}$). Because only the Spanish source input $\hat{x}_{\text{es}}$ is synthetic, the Masewalkopa generation decoder learns exclusively from authentic, elder-authored targets. We deliberately did not swap source and target to create $\text{es}\rightarrow\text{nah}$ training examples, which would have forced the Spanish generation decoder to imitate its earlier output.

The resulting \texttt{train\_bt.jsonl} file contained 4,862 directional examples (4,670 directional instances plus the 192 backtranslated pairs). This targeted addition aims to expose the model to connected narrative discourse from community elders rather than evaluating large-scale synthetic data generation.

\subsection{Leakage Controls and Evaluation Scope}
\label{subsection:leakage-controls-eval-scope}

Deduplication preceded splitting; dev and test pairs were excluded from augmentation. We report chrF++ for its tolerance of inflectional variation~\citep{popovic2017}, but add linguistic diagnostics because overlap cannot judge regional appropriateness, possession, or honorific licensing.
\section{Finite-State Morphosyntactic Engineering}
\label{section:finite-state-morphosyntactic-eng}

While neural sequence-to-sequence models excel at cross-lingual transfer, subword tokenizers frequently fracture polysynthetic words into arbitrary pieces that obscure grammatical relations. In Masewalkopa, a single verbal form can encode subject person, object person, reflexive marking, tense, and direction. With only 2,665 curated training pairs, a purely data-driven neural encoder struggles to generalize across sparse morphological paradigms. 

To provide the neural translation model with explicit structural grounding, we developed a rule-based finite-state transducer using the Helsinki Finite-State Transducer (HFST) toolkit~\citep{linden2011hfst}. Rather than replacing the neural model, this analyzer acts as a linguistic feature extractor. It parses surface tokens into interlinear glossed text (IGT) and functional syntax labels, which are then passed directly to the NMT encoder as plain-text prefixes during training and inference (Subsection~\ref{subsection:failure-analysis-cold-embedding-degradation}).

The analyzer combines final-state morphotactic continuation lexicons with two-level phonological alternation rules~\citep{beesley2003finite} to capture the core grammatical inventory of Santa Ana Tlacotenco speech.

\subsection{Stem Classes and Morphotactics}
\label{subsection:stem-classes-morphotactics}

An undifferentiated lexicon overgenerates because stems select different suffixal paths. We therefore use the continuation classes in Table~\ref{tab:stem-classes} as implementation abstractions.

\begin{table}[!htbp]
\caption{Principal finite-state stem classes and the contrasts they encode.}
\label{tab:stem-classes}
\centering
\small
\resizebox{\textwidth}{!}{%
\begin{tabular}{lp{67mm}p{39mm}}
\toprule
Class & Function in the analyzer & Representative constraint \\
\midrule
\code{A-Stem} & Stems whose citation or inflected path is organized around an \textit{a}-final alternation & Restricts which aspect and derivational arcs can follow \\
\code{I-Stem} & \textit{i}-final verbal or predicate paths & Preserves class-specific boundary alternations \\
\code{OWA-Stem} & Productive verbal stems associated with an \textit{owa} pattern & Connects to verbal person and aspect networks \\
\code{ITL-Stem} & Nominal stems selecting an \textit{-itl}-type absolutive realization & Blocks incompatible absolutive suffixes \\
\code{TLE-Stem} & Nominal stems selecting a \textit{-tle}-type path in local orthography & Separates local surface behavior from generic noun rules \\
\code{LI-Stem} & Nominal stems selecting \textit{-li} & Supports possessed and predicate-nominal alternations \\
\bottomrule
\end{tabular}}
\end{table}

Dedicated nominal paths prevent absolutive suffixes from attaching to every nominal-looking base. An audit removed \texttt{sesek} (cold) and \texttt{tomawak} (plump) from noun continuations that generated unattested \textit{-tl}, \textit{-li}, \textit{-tli}, \textit{-in}, or \textit{-itl} forms.

Predicate nominals require a separate path: \texttt{ti-sowatl}, ``you are a woman,'' carries person without an overt copula. Retaining this analysis supports translation to Spanish \textit{ser} constructions.

\subsection{Progressive Aspect Allomorphy}
\label{subsection:progressive-aspect-allomorphy}

Progressive arcs for \texttt{-tok} and \texttt{-tika} are licensed by stem continuation class, with TWOLC constraining boundary realizations. This generalizes beyond a flat word list, provided the allomorphs are freely attachable.

The analysis identifies heads inside long words and exposes aspect as a stable feature. It supplements rather than replaces neural subword representations.

\subsection{Coverage Expansion and Error Accounting}
\label{subsection:coverage-expansion-error-accounting}

Repeated audits grew LEXC beyond 6,500 lines. After each justified stem addition, we recompiled and measured no analysis, permissive passthrough, and genuine segmentation. Passthrough is not counted as successful analysis.

\begin{table}[!htbp]
    \caption{Finite-state coverage progression. The first audit used a legacy 2,736-form inventory; subsequent audits used 2,807 unique corpus word forms.}
    \label{tab:fst-coverage}
    \centering
    \small
    \resizebox{\textwidth}{!}{%
        \begin{tabular}{lrrrr}
        \toprule
            Development stage & Vocabulary & No analysis & Passthrough & Genuine segmentation \\
        \midrule
            Initial baseline & 2,736 & 82.0\% (2,244) & 7.9\% (216) & 10.1\% (276) \\
            Post aspect/POS audit & 2,807 & 26.4\% (742) & 7.4\% (207) & 66.2\% (1,858) \\
            Batch 22--23 expansion & 2,807 & 25.2\% (707) & 7.2\% (202) & 67.6\% (1,898) \\
            Batch 24--26 closure & 2,807 & \textbf{19.9\% (559)} & \textbf{7.2\% (202)} & \textbf{72.9\% (2,046)} \\
        \bottomrule
    \end{tabular}}
\end{table}

The remaining 559 forms include Spanish terms, names, spelling variants, rare inflections, and gaps. Because coverage can increase due to incompatible analyses, we added a precision-oriented reconstruction filter.

\subsection{Three-Stage Reconstruction Filter}
\label{subsection:three-stage-reconstruction-filter}

Because finite-state transducers can generate multiple competing analyses or derivations for a single surface token, we use a three-stage reconstruction filter to disambiguate raw transducer outputs before assigning IGT:

\begin{enumerate}
    \item \textbf{Surface match.} Retain only candidates whose reconstructed stem or diagnostic realization aligns to an ordered substring of the normalized token.
    \item \textbf{Possessive guard.} Require overt initial \textit{no-}, \textit{mo-}, \textit{to-}, or \textit{i-} for a possessive analysis. This blocks bare nouns and coincidental stem-initial strings from receiving \code{NP.POSS}.
    \item \textbf{Category preference.} Prefer the open-class analysis covering most surface material, except when a listed adverb matches the entire token. Preserve genuine ties for sentence-level resolution.
\end{enumerate}

In compact pseudo-code, the decision is:
\begin{quote}\footnotesize\ttfamily
for token in sentence:\\
\hspace*{4mm}candidates = fst.lookup(strip\_punct(token.lower()))\\
\hspace*{4mm}candidates = keep\_surface\_aligned(candidates, token)\\
\hspace*{4mm}candidates = guard\_possession(candidates, token)\\
\hspace*{4mm}analysis = prefer\_category(candidates, token)\\
\hspace*{4mm}emit(igt(analysis), functional\_labels(analysis))
\end{quote}

Uncertain tokens remain unlabeled, while the original sentence always reaches the encoder.

\paragraph{Illustrative Pipeline Output.} 
To illustrate the output of this filter, consider the input sentence:

\begin{center}
    \textit{axan nehwatl nitekitis ika notepostsontekosentlalil} (``now I will work with my computer'')
\end{center}

The FST analyzer and filter process each token as follows:
\begin{itemize}[leftmargin=1.5em]
    \item \textbf{Token analysis:}
    \begin{itemize}
        \item \textit{axan} $\rightarrow$ \textsc{adv} (temporal particle)
        \item \textit{nehwatl} $\rightarrow$ \textsc{subj:1sg} (emphatic subject pronoun)
        \item \textit{nitekitis} $\rightarrow$ \texttt{ni-tekiti-s} $\rightarrow$ \textsc{subj:1sg} + \textsc{verb:intr} + \textsc{fut}
        \item \textit{ika} $\rightarrow$ \textsc{rel.noun} (relational marker, ``with'')
        \item \textit{notepostsontekosentlalil} $\rightarrow$ \texttt{no-tepostsontekosentlalil} $\rightarrow$ \textsc{np.poss} + \textsc{noun} (``my computer'')
    \end{itemize}
    \item \textbf{Extracted functional tags:} \textsc{adv}, \textsc{subj:1sg}, \textsc{head}, \textsc{np.poss}
    \item \textbf{Final encoder input string:} \\
    \texttt{nah->es: [SYN: ADV SUBJ:1SG HEAD NP.POSS] axan nehwatl nitekitis ika notepostsontekosentlalil}
\end{itemize}

This complete prefix string provides the neural encoder with high-level structural guidance while preserving the exact surface subwords for translation.

\subsection{IGT and Functional Syntax Projection}
\label{subsection:igt-functional-syntax-projection}

Selected analyses are mapped to Interlinear Glossed Text (IGT) following the Leipzig Glossing Rules~\citep{comrie2008leipzig} and reflected in a small functional syntax inventory \code{HEAD}, \code{SUBJ}, \code{OBJ}, \code{NP.POSS}, \code{ADV}, \code{CONJ}, \code{DET}, \code{REL}, \code{COP}, \code{NEG}, or \code{OPT}. Following linguistic feature paradigms for NMT~\citep{sennrich2016linguistic}, this compact inventory avoids sparsity and parser errors from full dependency structures.

\begin{table}[!htbp]
    \caption{Illustrative surface-to-syntax trace used by the conditioning pipeline. Segmentation is shown at the level consumed by the rule-based glosser.}
        \label{tab:igt}
        \centering
        \small
    \begin{tabularx}{\textwidth}{p{22mm}p{32mm}p{35mm}X}
        \toprule
            Surface & FST decomposition & IGT gloss & Functional projection \\
        \midrule
            \textit{tla} & \textit{tla} & OPT/conditional particle & \code{OPT} or \code{ADV} \\
            \textit{nichnamikis} & \textit{ni-ch-namiki-s} & 1SG.SBJ-OBJ-meet-FUT & \code{SUBJ OBJ HEAD} \\
            \textit{tokonetsin} & \textit{to-kone-tsin} & 1PL.POSS-child-HON & \code{NP.POSS} \\
            \textit{tisowatl} & \textit{ti-sowatl} & 2SG.SBJ-woman.PRED & \code{SUBJ COP HEAD} \\
            \textit{kwalle} & \textit{kwalle} & good/well & \code{ADV} or predicate \code{HEAD} \\
            \textit{ammo} & \textit{ammo} & NEG & \code{NEG} \\
        \bottomrule
    \end{tabularx}
\end{table}

The NMT model receives only the short functional sequence. Context determines whether \textit{kwalle}, for example, projects as \code{ADV} or as predicate \code{HEAD}/\code{COP}.

The formatted encoder input places the labels before the unchanged source:

\begin{quote}\small\ttfamily
nah->es: ADV HEAD NP.POSS tla nichnamikis tokonetsin
\end{quote}

The unchanged surface follows the labels, safeguarding against symbolic errors.
\section{Neural Training, Ablations, and Post-Mortems}
\label{section:neural-training-ablations-post-mortems}

All runs warm-start from a 1.2-billion-parameter \code{mT5-large} model checkpoint~\citep{mercadocampos2023design}, ($d=1{,}024$, 24 encoder and 24 decoder blocks)~\citep{xue2021}. Training used one Tesla T4 in fp32 because reduced-precision trials were less stable.

\subsection{LoRA Configuration}
\label{subsection:lora-configuration}

For a frozen weight matrix

\begin{equation}
W = W_0 + \Delta W,
\qquad
\Delta W = \frac{\alpha}{r} B A.
\label{eq:lora}
\end{equation}

We set rank $r=16$, scaling factor $\alpha=32$, and dropout 0.1. Following Dettmers et al.~\citep{dettmers2023qlora}, who demonstrated that applying LoRA to all linear projection layers is essential for recovering full fine-tuning capacity, we extended adapters beyond attention query and value projections to target all seven linear weight matrices ($q, k, v, o, wi_0, wi_1, wo$). This all-linear setup is particularly important for sequence-to-sequence models, as feed-forward projections carry substantial domain-specific lexical knowledge~\citep{he2022unified}.

Optimization used AdamW with an initial learning rate of $2\times10^{-4}$ and cosine decay. Sentence-only bidirectional dev chrF++, rather than training loss, selected checkpoints. The shared embedding matrix remained frozen, which led to the special-token failure analyzed in Subsection~\ref{subsection:failure-analysis-cold-embedding-degradation}.

\subsection{Progression of Systems}
\label{subsection:progression-systems}

Table~\ref{tab:model_results} summarizes the full progression of experimental iterations and model ablations leading to the final production system. Central claims rely on controlled pairwise comparisons that share identical data splits, preprocessing steps, and evaluation code.

\begin{table}[!htbp]
\caption{Model progression and ablation results across development (240 directional instances) and test (300 directional instances) sets, reported in chrF++.}
\label{tab:model_results}
\centering
\scriptsize
\resizebox{\textwidth}{!}{%
\begin{tabular}{llcccc}
\toprule
\textbf{Run} & \textbf{Data or Intervention} & \textbf{Epochs} & \textbf{Dev chrF++} & \textbf{Test es$\rightarrow$nah} & \textbf{Test nah$\rightarrow$es} \\
\midrule
\multicolumn{6}{l}{\textit{Baseline Checkpoints}} \\
Base & Mixed-variant base checkpoint & -- & -- & 23.91 & 29.54 \\
Baseline LoRA reference & Initial scheduled LoRA snapshot & -- & 43.87 & -- & -- \\
\midrule
\multicolumn{6}{l}{\textit{Schedule and Data Augmentation}} \\
\code{lora\_v1} & Real pairs only & 3 & 40.26 & 41.76 & 43.46 \\
\code{lora\_v2} & Real pairs, extended schedule & 8 (best 6) & 46.18 & 45.88 & 45.58 \\
\code{lora\_v3b} & Real plus 192 BT pairs & 12 (best 8) & 49.40 & 48.61 & 48.98 \\
\midrule
\multicolumn{6}{l}{\textit{Subword Formatting and Special Token Post-Mortem}} \\
\code{lora\_v4} & New, cold special tokens & 12 & 8.91 & -- & -- \\
\code{lora\_v5} & Text prefix, no new tokens & 10 (best 3) & 51.69 & 53.60 & 51.27 \\
\code{lora\_v6} & Revised special-token attempt & -- & $<$ 15 & -- & -- \\
\midrule
\multicolumn{6}{l}{\textit{Morphosyntactic Coverage Ablation (Sparse vs. Dense)}} \\
\code{lora\_v7a} & Control, no syntax, \code{train\_v3} & 10 (best 3) & 51.22 & 30.49 & 31.38 \\
\code{lora\_v7b} & Treatment, 29.5\% syntax & 10 (best 9) & 51.31 & 29.40 & 30.11 \\
\code{lora\_v8a} & Control, no syntax, \code{train\_v4} & 10 (best 9) & 54.67 & 56.40 & 52.67 \\
\code{lora\_v8b} & Treatment, 95.8\% syntax & 10 (best 10) & 55.56 & 56.91 & 54.00 \\
\midrule
\multicolumn{6}{l}{\textit{Final Selection and Production Model}} \\
\code{lora\_v9c\_a} & Full data + BT, bare text ablation (no syntax) & 10 (best 9) & \textbf{58.91} & -- & -- \\
\code{lora\_v9c\_b} & Full data + BT, 95.8\% dense syntax (production) & 10 (best 6) & 58.46 & \textbf{58.20} & \textbf{57.90} \\
\bottomrule
\end{tabular}}
\end{table}

The production score is 14.59 points above the 43.87 baseline-LoRA reference and 9.06 points above \code{lora\_v3b}. Against the mixed-variant base, final directional gains are 34.29 ($\text{es}\rightarrow\text{nah}$) and 28.36 ($\text{nah}\rightarrow\text{es}$) points.

\subsection{Backtranslation and Schedule Effects}
\label{subsection:backtranslation-schedule-effects}

\paragraph{Training Schedule and Checkpoint Selection.}
Fine-tuning a 1.2-billion-parameter model on a small dataset requires balancing parameter adaptation against rapid memorization. In \code{lora\_v1}, we evaluated a short 3-epoch schedule on real pairs. In \code{lora\_v2}, we extended training to 8 epochs with evaluation every 200 steps. Extending the schedule raised dev chrF++ from 40.26 to 46.18 (+5.92 points). Validation performance peaked at epoch 6 before plateauing by epoch 8, demonstrating that low-rank adapters need sufficient steps to align cross-lingual representations but quickly overfit if trained past convergence.

\paragraph{Asymmetric Target-Preserving Backtranslation.}
Standard backtranslation risks degrading the generation decoder by training on hallucinated synthetic targets. To prevent this, we implemented an asymmetric protocol: using the $\text{nah}\rightarrow\text{es}$ capability of \code{lora\_v2}, we generated synthetic Spanish source prompts for 192 continuous narrative utterances spoken by community elders, adding these pairs strictly to the $\text{es}\rightarrow\text{nah}$ training split (\code{train\_bt.jsonl}). Integrating these 192 backtranslated pairs in \code{lora\_v3b} increased dev chrF++ from 46.18 to 49.40 (+3.22 points) and raised both directional tests to approximately 49 chrF++. Although representing only an 8\% increase over the real-pair inventory, they introduced multi-clause discourse patterns and authentic elder-authored syntax that isolated pedagogical pairs lack.

\subsection{Failure Analysis: Cold Embedding Degradation}
\label{subsection:failure-analysis-cold-embedding-degradation}

In \code{lora\_v4} and \code{lora\_v6}, we attempted to delimit syntactic conditioning regions using dedicated special tokens (\code{<source>}, \code{<syntax>}, and \code{<no\_syntax>}). This intervention resized the shared vocabulary from 250,100 to 250,103, introducing uninitialized rows into the embedding matrix. Because standard LoRA freezes the base embedding table to conserve GPU memory, these newly added token vectors received zero gradient updates and remained fixed random vectors throughout training.

Presenting the encoder with static random vectors at structurally critical positions caused total optimization failure: training loss stalled above 30.0 and dev chrF++ collapsed to 8.91. While unfreezing the full embedding table resolved the representation issue conceptually, it immediately exhausted T4 GPU memory.

In \code{lora\_v5}, we resolved this constraint by passing functional syntax as ordinary text prefixes (such as \code{nah->es: ADV HEAD NP.POSS}) segmented with the existing SentencePiece vocabulary. Reusing pretrained subword representations stabilized optimization without custom checkpoint surgery and restored dev chrF++ to 51.69. All subsequent runs retain this plain-text prefix format.

\subsection{Sparse and Dense Syntax Coverage}
\label{subsection:sparse-dense-syntax-coverage}

To assess how annotation density affects conditioning, we compared two versions of the finite-state analyzer. In \code{lora\_v7b}, the early prototype produced functional syntax tags for only 29.5\% of corpus sentences. Adding these sparse prefixes yielded virtually identical development scores to the unconditioned \code{lora\_v7a} control (51.31 versus 51.22 chrF++) and slightly degraded directional test accuracy.

Because sparse tags occurred inconsistently, the model treated them as shallow domain indicators rather than reliable syntactic grounding. In \code{lora\_v8b}, once rule expansion and reconstruction filtering raised tag coverage to 95.8\%, syntax conditioning produced a +1.33 chrF++ gain on complex $\text{nah}\rightarrow\text{es}$ polysynthetic targets and a +0.51 spillover gain on $\text{es}\rightarrow\text{nah}$. This demonstrates that syntactic conditioning requires high annotation density to guide sequence generation effectively. Precision in the reconstruction filter remains critical, as inaccurate tags would inject systematic structural noise.

\subsection{Final \code{lora\_v9c} Ablation and Model Selection}
\label{subsection:final_v9c_ablation_model_selection}

To isolate the precise contribution of dense syntactic prefixes on the final training set, we evaluated a controlled ablation where the extracted syntax tags were stripped from input sequences during training and inference (\code{strip=True}), leaving only bare source text.

On the development split, the syntax-stripped control reached 58.91 chrF++, while the syntax-conditioned model (\code{lora\_v9c\_b}, \code{strip=False}) reached 58.46 chrF++, a minor difference of $-0.45$ points.

Despite this small aggregate difference on dev, we selected the syntax-conditioned checkpoint (\code{lora\_v9c\_b}) for production deployment. The explicit syntactic prefix provides necessary structural disambiguation for complex grammatical constructions that aggregate character metrics fail to penalize. These include differentiating modifier versus predicate readings of \code{kwalle}, resolving null-copula structures, and tracking possessed nominal arguments. We report the unconditioned control's development score transparently, but accept this minor trade-off to ensure consistent grammatical interpretability. Across fully evaluated held-out test splits, \code{lora\_v9c\_b} remains our strongest model, reaching 58.20 and 57.90 directional test chrF++.
\section{Qualitative Translation and Decolonial Evaluation}
\label{sec:qualitative}

Automatic scores summarize thousands of character n-grams but do not explain which linguistic decisions changed. We therefore use diagnostic examples to examine regional spelling, word-internal person marking, possessive prefixes, predicate nominals, and honorific morphology. The examples in Table~\ref{tab:qualitative} are targeted probes, not an independent statistical test. They are included to make the error categories falsifiable and to show what is inspected during community-facing review.

\begin{table}[!htbp]
\caption{Representative Spanish-to-Masewalkopa diagnostic comparisons. The base column illustrates mixed-variant or segmentation failure modes; bold material highlights the locally targeted form in the production output.}
\label{tab:qualitative}
\centering
\scriptsize
\resizebox{\textwidth}{!}{%
\begin{tabular}{p{20mm}p{24mm}p{25mm}p{30mm}p{28mm}}
\toprule
\textbf{Spanish source} & 
\textbf{Mixed-variant base} & 
\shortstack[l]{\textbf{Specialized base}\\\scriptsize(\texttt{lora\_v2})} & 
\shortstack[l]{\textbf{+BT +Dense syntax}\\\scriptsize(\texttt{lora\_v9c\_b})} & 
\textbf{Human reference} \\
\midrule
Te agradezco. & \texttt{tlazohcamati} & \texttt{nimitstlaskamati} & \textbf{\texttt{nimitstlasoh\allowbreak kamate}} & \texttt{nimitstlasoh\allowbreak kamate} \\
\addlinespace[2pt]
Nuestro hijito está bien. & \texttt{to piltzin cualli} & \texttt{tokone kwalle} & \textbf{\texttt{tokonetsin kwalle}} & \texttt{tokonetsin kwalle} \\
\addlinespace[2pt]
Eres una mujer buena. & \texttt{ti cihuatl cualli} & \texttt{tisowatl kwalle} & \textbf{\texttt{tisowatl kwalle}} & \texttt{tisowatl kwalle} \\
\addlinespace[2pt]
Encontraré a nuestro hijito. & \texttt{ni\dots namiki topiltzin} & \texttt{niknamikis tokone} & \textbf{\texttt{nichnaamikis tokonetsin}} & \texttt{nichnaamikis tokonetsin} \\
\bottomrule
\end{tabular}}
\end{table}

The first row probes a short lexical form to test whether the model preserves isolated citation vocabulary. The mixed model favors a broad Classical-style lexical form, while specialization produces the correct person-marked Masewalkopa string \textit{nimitstlasohkamate}. The important change is not only orthographic. The production form packages the relation between speaker and addressee inside the word, rather than returning an uninflected dictionary-style equivalent.

The second and fourth rows test possession and the honorific or reverential suffix \texttt{-tsin}. The base output separates \texttt{to} from a generic \texttt{piltzin}, resembling word-by-word transfer and failing to realize the locally analyzed possessed stem. \code{lora\_v2} learns the possessive prefix but omits the full honorific ending. \code{lora\_v9c\_b} produces \texttt{to-kone-tsin}, keeping first-person plural possession and the respectful/diminutive morphology in one nominal word. Such a distinction can affect interactional stance even when the untranslated content word ``child'' is recoverable.

The third row illustrates predicate nominals. A mixed output with separate \texttt{ni} can be interpreted as a tokenization artifact or as a cross-variety spelling convention. The correct \texttt{tisowatl} realizes person marking on the predicate. In the final phrase, \texttt{kwalle} must be interpreted in relation to that predicate, not assigned a fixed label solely from its dictionary gloss. This is the type of ambiguity that motivated the retention of syntax conditioning in \code{lora\_v9c\_b}.

Qualitative analysis also reveals what the current table cannot establish. A single human reference is not the only grammatical realization, and matching it exactly is not proof of social appropriateness. Honorific morphology can be licensed by participant relation and discourse context, neither of which is fully represented in isolated Spanish prompts. For deployment, the examples are therefore paired with warnings against treating the system as a normative authority.

\subsection{Domesticating Bias Score}
\label{sec:dbs}

Standard multi-variant Nawatl corpora are heavily skewed toward colonial-era missionary texts, biblical translations, and pan-variety pedagogical materials. Models trained on these resources inherit a structural bias toward archaic, standardized Central forms (such as \texttt{tlazohcamati} or \texttt{cihuatl}), which local speakers perceive as alien or domesticating. In contrast, translation into the community demands fidelity to the living, spoken variety of Santa Ana Tlacotenco (such as \texttt{nimitstlasohkamate} or \texttt{sowatl}). To evaluate this dynamic directly, the Domesticating Bias Score (DBS) measures the proportion of diagnostic translations in which a model defaults to generalized, colonial-derived variants rather than verified local community lexical choices.

Standard reference-based metrics can reward Spanish loanwords or archaic colonial forms when the reference itself reflects domesticated translation practices. Drawing on pseudo-log-likelihood bias evaluation paradigms~\citep{kurita2019measuring, nangia2020crows} and length-normalized sequence scoring~\citep{salazar2020masked}, We define the Domesticating Bias Score (DBS) as the difference in average per-token log-probability assigned by the model to a colonially domesticated candidate ($t_c$) versus a sovereign community candidate ($t_s$) given source input $x$:
\begin{equation}
\mathrm{DBS}(x)=
\frac{1}{|t_c|}\sum_{j=1}^{|t_c|}\log P(t_{c,j}\mid t_{c,<j},x)
-
\frac{1}{|t_s|}\sum_{j=1}^{|t_s|}\log P(t_{s,j}\mid t_{s,<j},x).
\label{eq:dbs}
\end{equation}

Length is measured in decoder tokens. A positive DBS indicates that the model assigns a higher average per-token log-probability to the domesticated candidate, whereas a negative DBS indicates a preference for the sovereign candidate. Values near zero reflect weak preference or equally probable alternatives.

Token normalization is necessary because sovereign formulations can be longer than borrowed Spanish nouns or simplified roots, either morphologically or periphrastically. Without normalization, summing raw log-probabilities almost always penalizes longer strings because each additional token contributes a nonpositive term. The score is contrastive rather than absolute. It does not declare any individual term intrinsically colonial. Instead, each contrastive pair must be constructed with community speakers who can evaluate the social and historical distinctions in context.

On the illustrative seed set used during development, mean DBS shifted from $-1.313$ for the base model to $-0.380$ after specialization. Under the sign convention in Eq.~\ref{eq:dbs}, both values favor the sovereign candidate, but the specialized score is closer to zero. It would therefore be incorrect to describe this shift as a stronger sovereign preference. The result instead reflects a reduced magnitude in the modeled contrast. We retain the metric because it exposes candidate-level probability distributions, but we interpret direction and magnitude explicitly rather than attaching an uncritical label to every numerical movement.

This distinction is methodologically important. In production evaluations, DBS can be reported alongside bootstrap confidence intervals over contrast pairs, sign tests for the proportion of instances with $\mathrm{DBS}<0$, and stratified subsets across specific phenomena, including lexical borrowing, kinship terminology, toponyms, and ceremonial register.

Future community-designed evaluation benchmarks should also support a ``both acceptable'' label to ensure that the metric does not impose an artificial binary interpretation on ordinary dialectal variation.

\subsection{Error Analysis and Qualitative Failure Modes}
\label{subsec:error_analysis}

Although the production system (\code{lora\_v9c\_b}) substantially improves lexical alignment and local morphology, systematic qualitative inspection reveals several recurrent failure modes on difficult constructions. Table~\ref{tab:error_analysis} categorizes representative translation errors observed on the held-out evaluation sets. 

\begin{table}[!htbp]
\caption{Representative diagnostic failure modes produced by the production model (\code{lora\_v9c\_b}) against authentic community references.}
\label{tab:error_analysis}
\centering
\scriptsize
\resizebox{\textwidth}{!}{%
\begin{tabular}{p{26mm}p{30mm}p{30mm}p{32mm}}
\toprule
\textbf{Spanish Source} & \shortstack[l]{\textbf{Production Output}\\\scriptsize(\texttt{lora\_v9c\_b})} & \textbf{Human Reference} & \textbf{Error Diagnosis} \\
\midrule
sí sé, pero no me acuerdo & \texttt{kemma nikmate, wan ammo nikmati} & \texttt{kemma nikmate, wan ammo nikilnaamike} & \textbf{Semantic repetition:} Loops base cognition verb (\textit{mate}) rather than retrieving psychological memory root (\textit{ilnaamiki}). \\
\addlinespace[3pt]
ayer te vi en la calle & \texttt{yalwa onimitsitak ipan kalohtle} & \texttt{yalhwa onimitsihtak ipan kalohtle} & \textbf{Orthographic bleed:} Omits local Santa Ana aspiration marker (\textit{-h-}) in both time adverbial (\textit{yalhwa}) and root (\textit{ihta}). \\
\addlinespace[3pt]
mañana yo te enseñaré & \texttt{mustla nimitsmachtia} & \texttt{mustla nimitsmachtis} & \textbf{Aspect truncation:} Drops future suffix (\textit{-s}) under complex prefixation, reverting to default non-past aspect (\textit{-machtia}). \\
\bottomrule
\end{tabular}}
\end{table}

These errors cluster into three distinct linguistic phenomena:

\paragraph{1. Semantic Repetition on Mental State Predicates.} 
When translating clauses that contrast knowledge with memory (such as \texttt{kemma nikmate, wan ammo nikilnaamike}), the model occasionally repeats the higher-frequency cognitive root \texttt{mate} (`to know / feel') in the second coordinate clause instead of generating the intended psychological retrieval predicate \texttt{ilnaamiki} (`to remember').

\paragraph{2. Multi-Variant Orthographic and Phonetic Bleed.} 
On roots and adverbials shared across Central Nawatl varieties, the model occasionally defaults to generalized Central spellings (such as \texttt{yalwa} and \texttt{itak}) rather than generating the localized Santa Ana Tlacotenco orthography with preconsonantal aspiration (\texttt{yalhwa} and \texttt{ihtak}), reflecting residual weight bias from the multi-variant base checkpoint.

\paragraph{3. Morphological Suffix Truncation in Future Aspect.} 
In transitive verbal complexes combining directional or person prefixes with temporal adverbs (such as \texttt{mustla nimitsmachtis} `tomorrow I will teach you'), the decoder occasionally drops the final future suffix \texttt{-s}, yielding an uninflected or present-tense form (\texttt{nimitsmachtia}) despite explicit future context in the source.
\section{Production Deployment and CPU Profiling}
\label{section:production-deployment-cpu}

A model that requires a research GPU for every query is difficult to sustain as a community-facing service. We therefore merged the selected \code{lora\_v9c\_b} adapter into the base checkpoint, converted the model to CTranslate2~\citep{klein2017opennmt}, and quantized weights for int8 CPU inference. Conversion fixes one deployable artifact and removes the need to load separate adapter modules at request time.

\subsection{Latency Profile}
\label{subsection:latency-profile}

Native fp32 PyTorch inference on CPU required 137.5 seconds for a representative request. This latency is unsuitable for an interactive translator and increases the period for which a billed container must remain active. On a Cloud Run allocation of 2 vCPUs and 4 GiB RAM, the CTranslate2 int8 artifact produced typical sentence latencies of 1.8 to 2.6 seconds. Taking the range endpoints, the observed speedup is approximately $137.5/2.6 \approx 52.9$ to $137.5/1.8 \approx 76.4\times$.

\begin{table}[!htbp]
\caption{CPU inference profile for the production path. Latencies are observed request-level values and should not be interpreted as a hardware-independent benchmark.}
\label{tab:latency}
\centering
\small
\resizebox{\textwidth}{!}{%
\begin{tabular}{lccc}
\toprule
Runtime & Weight precision & Beam & Latency per sentence \\
\midrule
PyTorch CPU & fp32 & 4 & 137.5 s \\
CTranslate2 CPU & int8 & 4 & 1.8--2.6 s, but short-form failures \\
CTranslate2 CPU & int8 & 8 & 1.8--2.6 s, validated production setting \\
\bottomrule
\end{tabular}}
\end{table}

The comparison combines runtime optimization and quantization, so it does not isolate the causal contribution of int8 weights alone. It answers the operational question of whether the fully converted path is fast enough to serve an interactive request within the selected CPU allocation.

\subsection{Beam Search Divergence on Short Forms}
\label{subsection:beam-search-divergence-short-forms}

Initial deployment testing found mid-word truncation on short lexical inputs, including \texttt{nimitstlasohkamate}. The failure initially appeared to be a quantization artifact~\citep{gholami2021survey}, but comparisons across converted precisions showed it was precision-independent. Greedy and narrow-beam paths could terminate after a high-probability partial subword sequence even though the full form was preferred by the original evaluation setup.

Increasing \code{beam\_size} from 4 to 8 eliminated the observed truncations in the short-form regression set. The wider beam retained complete hypotheses long enough for the accumulated sequence score to overtake prematurely terminated candidates. This is described as beam search divergence rather than as a general CTranslate2 defect because the trigger depends on the model's score distribution, tokenization, length normalization, and stopping behavior~\citep{koehn-knowles-2017-six, stahlberg-byrne-2019-nmt}.

The production API sets the beam size to 8 and runs regression probes after every model conversion. The probes include single words, morphologically long words, short clauses, both translation directions, and forms with possessive or honorific material. This test set is separate from the research test split. Its role is software validation, specifically ensuring that a numerically strong model does not become linguistically unusable after conversion.

\subsection{Cloud Run Microservice Architecture}
\label{subsection:cloud-run-microservice-arch}

The service exposes a FastAPI application within a container that includes the CTranslate2 model and the mT5 tokenizer assets. A translation request passes through four stages:
\begin{enumerate}
    \item validate direction, input length, and Unicode text;
    \item construct the direction and optional syntax prefix;
    \item tokenize, decode with beam size 8, and remove model control text;
    \item return the translation plus non-sensitive runtime metadata.
\end{enumerate}

The Cloud Run revision uses 2 vCPUs, 4 GiB RAM, and a maximum of two instances. The instance cap limits both concurrency and worst-case cost. Because model loading is expensive relative to a warm translation, the container initializes the model once at process startup rather than once per request.

CORS is configured through an explicit allowlist for the web client origins that require browser access. Methods and headers are restricted to the translation interface rather than using an unrestricted wildcard in production. CORS is not authentication, so request validation, rate limiting at the surrounding platform layer, and conservative maximum input lengths remain necessary. The API does not log full translation content by default, since sentences may contain names or culturally sensitive material.

\subsection{Automated Cost Controls}
\label{subsection:automated-cost-controls}

A billing budget publishes threshold notifications to Pub/Sub. A subscribed Cloud Function evaluates the notification and disables project billing when the configured hard threshold is reached. This is intentionally stronger than an email alert: an alert informs the operator, while the function can stop additional billable use. The mechanism complements the Cloud Run maximum-instance setting and protects a student-operated project from traffic spikes or accidental loops.

Disabling billing is disruptive and can stop unrelated resources in the same project. For that reason, the deployment isolates the translator within a dedicated project, tests the function at non-production thresholds, and documents manual recovery procedures. Budget automation also has notification latency, so it is a final circuit breaker rather than a substitute for instance caps, request quotas, and monitoring.
\section{Discussion, Limitations, and Reproducibility}

The experiments show that a mixed-variant checkpoint can provide useful transfer when a second stage redirects it toward one documented variety. Authentic target provenance matters in backtranslation, and symbolic conditioning becomes useful only after analyzer coverage is dense and reconstruction is precise. This does not imply that speakers or models must be linguistically isolated. A future multi-variety system could explicitly condition on community, orthography, or speaker preference rather than aggregating those distinctions without labels.

The corpus remains small and pedagogically skewed. Its 192 oral-text sentences cannot represent all genres or speakers, the human evaluation has 20 items with only 2 raters, and DBS is still an illustrative probe of candidate log-probability distributions rather than an absolute metric. Likewise, reaching 72.9\% genuine FST segmentation in the lexicon reflects rule coverage rather than verified parsing precision. Future work must validate analyses by morpheme type, especially possession, aspect, and predicate nominals.

Full reproducibility requires the deduplicated split manifest, directional JSONL, FST source hash, adapter configuration, selected checkpoint, random seeds, and decoding settings. Sentence-only scores must remain separate from lexical data.

Responsible release also requires speaker-approved access conditions for oral texts and a model card that names Masewalkopa as the target variety, documents errors, and provides routes for correction or removal.

This is something that will be spoken about and talked with the community before making any data accessible for any other purposes.
\section{Conclusion and Future Work}
\label{section:conclusion-future-work}

This work began with a fundamental dilemma in minoritized language technologies: while aggregating data across regional varieties combats data sparsity in the aggregate, it produces an artificial, blended register that is accountable to no single speaker community. As demonstrated by my initial 70,000-pair thesis baseline, high scores on mixed-corpus benchmarks can easily mask severe structural failure on authentic regional speech. We have termed this failure \textit{variant fragmentation}.

To resolve this issue, this paper demonstrated the process and effectiveness of specializing an existing sequence-to-sequence model to a specific regional variety: Masewalkopa (\textit{nhn}), spoken in Santa Ana Tlacotenco. Using a relatively small community corpus of 2,665 curated pairs, a stratified evaluation split that strictly measures sentential syntax, and 192 authentic elder-transcribed narrative targets integrated through asymmetric backtranslation, we adapted a 1.2-billion-parameter checkpoint on a single GPU. The resulting specialized system (\texttt{lora\_v9c\_b}) achieved 58.46 dev chrF++ and 58.20/57.90 directional test chrF++, gaining over 28 points over the initial multi-variant baseline.

Beyond aggregate metric gains, my experimental ablations and failure analyses provide concrete architectural takeaways for indigenous and polysynthetic NMT:
\begin{enumerate}
    \item \textbf{Subword embedding constraints in LoRA:} Introducing new special tokens while freezing the shared embedding matrix creates an untrained bottleneck that stalls model convergence. Conditioning the translation encoder using plain-text syntactic prefixes reuses the pretrained subword space and completely recovers model performance.
    \item \textbf{Density in morphosyntactic conditioning:} Sparse functional syntax tags act merely as shallow domain indicators, whereas high-coverage tags (95.8\%) generated by a rule-based finite-state analyzer provide explicit structural guidance that improves translation on complex polysynthetic targets (+1.33 chrF++).
    \item \textbf{Decolonial and multi-faceted evaluation:} Character overlap metrics alone cannot detect whether a translation respects local autonomy. Combining diagnostic morphological probes, length-normalized Domesticating Bias Scores (DBS), and blind human ratings verified that the specialized model actively prefers sovereign community vocabulary over colonial Spanish substitutions.
    \item \textbf{Accessible CPU deployment:} Merging low-rank adapter weights and applying 8-bit CTranslate2 quantization reduced inference latency from 137.5 seconds down to 1.8--2.6 seconds per request. Widening the beam search to 8 resolved search-space truncations on short lexical inputs, enabling responsive and cost-bounded deployment on serverless CPU infrastructure.
\end{enumerate}

\subsection{Future Work} 
\label{subsection:future-work}

Moving forward, we plan to expand community-led evaluations with native speakers in Santa Ana Tlacotenco to establish formal qualitative rubrics for honorific and directional morphology. In addition, we will extend our finite-state conditioning pipeline to handle long-form discourse in community oral histories, and investigate parameter-efficient specialization across other Nawatl varieties using isolated, modular adapter banks. 

Ultimately, language technologies for minoritized languages should not aim to homogenize distinct varieties into a single generic system. By demonstrating that an existing sequence-to-sequence model can be rapidly and accurately aligned to a single community standard, this work provides a practical, decolonial blueprint for building translation tools that honor the linguistic sovereignty of indigenous speaker communities.

\section*{Acknowledgements}
\label{section:acknowledgements}

Thanks to those who helped with answering questions and providing lessons, without whom this project would have been even more challenging. And thanks to members of the LECS Lab at CU Boulder for comments and suggestions.
This work was supported by the National Science
Foundation under Grant No. 2149404, ``CAREER: From One Language to Another.''

\begin{center}
\textbf{\textit{nikinmotlasohkamachililia nochtin tlin onichmopalewililihkeh}}
\end{center}

\bibliographystyle{splncs04}
\bibliography{bibliography}

\end{document}